\pdfoutput=1

\documentclass[11pt]{article}

\usepackage{acl}

\usepackage{times}
\usepackage{latexsym}

\usepackage[T1]{fontenc}

\usepackage[utf8]{inputenc}

\usepackage{microtype}

\usepackage{amsmath}
\usepackage{graphicx}
\usepackage{MnSymbol}%
\usepackage{wasysym}%

\title{To Know by the Company Words Keep and What Else Lies in the Vicinity}

\author{Jake Ryland Williams \\
  Drexel University \\
  Department of Information Science \\
  \texttt{jw3477@drexel.edu} \\\And
  Hunter Scott Heidenreich \\
  Harvard University \\
  Department of Computer Science \\
  \texttt{hheidenreich@g.harvard.edu} \\}

\begin{document}
\maketitle
\begin{abstract}
The development of state-of-the-art (SOTA) Natural Language Processing (NLP) systems has steadily been establishing new techniques to absorb the statistics of linguistic data. These techniques often trace well-known constructs from traditional theories, and we study these connections to close gaps around key NLP methods as a means to orient future work. For this, we introduce an analytic model of the statistics learned by seminal algorithms (including GloVe and Word2Vec), and derive insights for systems that use these algorithms and the statistics of co-occurrence, in general. In this work, we derive---to the best of our knowledge---the first known solution to Word2Vec's softmax-optimized, skip-gram algorithm. This result presents exciting potential for future development as a direct solution to a deep learning (DL) language model's (LM's) matrix factorization. However, we use the solution to demonstrate a seemingly-universal existence of a property that word vectors exhibit and which allows for the prophylactic discernment of biases in data---prior to their absorption by DL models. To qualify our work, we conduct an analysis of independence, i.e., on the density of statistical dependencies in co-occurrence models, which in turn renders insights on the distributional hypothesis' partial fulfillment by co-occurrence statistics. 
\end{abstract}

\section{Motivation} \label{sec:motivation}
Suppose one wished to randomly optimize a Rube Goldberg machine (RGM) over many Dominoes with the intent of accomplishing a small downstream task. Should the RGM be initialized to a random state, with dominoes scattered haphazardly, i.e., with no prior? Or would it help more to constrain the RGM to initializations with all dominoes standing on end? Perhaps less effort could be used to modify the dominoes-on-end state for the goal---but that depends on the goal and how dominoes can be used to transfer energy over long ranges. 

Pre-trained models are often used as \textit{initializations}, eventually applied to downstream NLP tasks like part-of-speech tagging or machine translation. This means model pre-training is a lot like initializing an RGM to a highly-potentiated state, while retaining a flexibility/generality to optimize sharply for the diversity of phenomena which can depend on statistical, linguistic information. A challenge partly met by big data pre-training is with the need for models to remain useful on a large diversity of data and tasks. Under the RGM theory, pre-training over big data simply potentiates more dominoes, in more-usefully correlated ways, where `useful' is hands-off defined by a model's parametric ability to explain language, i.e., which words were where. However, if we knew how many dominoes should be on end at the start and how many dominoes should be in configurations that make stairs, etc., it seems plausible to initialize the RGM with distributionally-useful tools, given what we know about how humans use dominoes to transfer energy, i.e., the statistics of how humans use vocabularies to communicate. We investigate these questions, replacing `domino' with `parameter', and lay the groundwork for provisioning statistical priors to efficiently meet model pre-training needs for future research, while uncovering a cost-effective method for probing the biases that DL models will learn if they train on specific data.

\section{Introduction} \label{sec:introduction}
GPT-3 is an off-the-shelf AI that is perhaps the pinnacle of LMs, and compared to GPT-2, was basically just trained on more data and with more parameters. The data that trained GPT-3's SOTA unsupervised machine translation performance were simply an ``unfiltered distribution of languages reflected in internet text datasets'' \cite{brown2020}. To uncover how a blend of training data like this aligns the semantics of, e.g., French and English vocabularies requires explaining \textit{what} (statistically) NLP tasks are teaching to SOTA algorithms, e.g., via the language modeling or masked language modeling tasks. However, while off-the-shelf AIs like GPT-2 are becoming ubiquitous in applications, they're also being shown to contain dangerous biases that emerge from training data~\cite{wallace19,heidenreich2021}.

\subsection{Related Work} \label{sec:related-work}
Within the last decade, there have been major shifts in representation learning from context-independent word vectors \cite{mikolov2013efficient,mikolov2013distributed,pennington2014glove}, to context-dependent word representations \cite{howard2018universal,peters2018deep}, to pre-trained language models \cite{devlin2019bert,radford2018improving,radford2019language}. These trends have been accompanied by large architectural developments from the dominance of RNNs \cite{hochreiter1997lstm}, to the appearance of attention \cite{bahdanau2015neural} and the proliferation of the Transformer architecture \cite{vaswani2017attention}. Our work seeks to open a path towards the \textit{efficient} engineering of SOTA NLP technologies. We aim to compute the natural statistics to which model parameters converge, and towards this our work analyzes the older, static-representations that preceded large LMs.

Despite gains on empirical benchmarks, recent works suggest surprising findings: word order may not matter as much in pre-training as previously thought \cite{sinha2021masked}, random sentence encodings are surprisingly powerful \cite{wieting2018no}, one can replace self-attention operations in BERT \cite{devlin2019bert} with unparameterized Fourier transformations and still retain 92\% of the original accuracy on GLUE \cite{lee2021fnet}, and many modifications to the Transformer architecture do not significantly impact model performance \cite{narang2021transformer}. There's no denying increases in empirical performance, but these confounding results raise questions about these models and the processing needed to perform NLP tasks.

\section{Harmonically-Distributed Data} \label{sec:harmonic-frequency}
Historically, research has naïvely approached the characterization of language statistics by counting the occurrence of symbols. While \textit{occurrence frequency} can be measured at different levels, e.g., characters, tokens, or phrases, a statistical ubiquity was discovered early on for tokens---specifically the harmonic relationship which exists in the usage of a document's vocabulary~\cite{zipf1935a,zipf1949a}. To understand the harmonic relationship, suppose a vocabulary $\mathcal{V}$ of $|\mathcal{V}| = N$ distinct \textit{types} is used to convey a collection of documents, $\mathcal{D}$, containing $M$ tokens. A harmonic analysis of $\mathcal{D}$ first \textit{ranks} each $t\in \mathcal{V}$ with a positive integer $r_t$ that sorts the vocabulary from high-to-low by frequency. Intuitively a rank, $r_t$ indicates the number of \textit{other} types which occur at least as often as $t$ (without loss of generality). Via this ranking, the empirical occurrence frequency for any type, $f_t$, can be mathematically approximated by harmonically-proportioned values: $f_t\approx N\cdot r_t^{-1}$, where $N$ scales models to have least-frequent types occur once. Crudely, harmonic distributions describe the bulk of statistical structure in token-frequency distributions.

\subsection{Co-Occurrence and Context}
Token co-occurrence matrices, i.e., co-frequency distributions, measure the number of times tokens appear, but specifically, `near' one another. In general, for types $t$ and $s$, we denote the occurrence of $s$ in a fixed window of size $\pm m$ tokens around $t$ across a collection, $\mathcal{D}$, by $F^m_{t,s}$. Most of the seminal representation-learning algorithms (including LSA, Word2Vec, and GloVe) rely on such empirical, $m$-sampled `data' of co-occurrence. Here, it's important to note that totality in co-occurrence distributions is dependent on the size of the context window, i.e., co-occurrence marginalization, which we denote by $M_F^m$, exhibits how the distribution `inflates' with larger values of $m$: $M_F^m = \sum_{t,s\in\mathcal{V}}F^m_{t,s} = \mathcal{O}\left(2mM\right)$. This $m$-window inflation thus slightly re-defines unigram statistics along marginals, denoted: $f^m_t = \sum_{l\in V} F^m_{t,l}$.

Generally, word co-occurrences define a specific family of word-context joint-distributional models, or, \textit{context distributions}, which can be tuned, e.g., to count only forward, backward, or any un-centered `windows' of context. These can likewise be generalized to $n$-gram context models~\cite{piantados2011}. While the over-counting effects of co-occurrence and $n$-gram contexts can be alleviated to form integrated higher-order models via weighted context distributions~\cite{williams2015c}, no representations have to-date used these models. Here, our work is again retrospective, focusing on building solid foundations from the standard, symmetrically-centered word co-occurrence model of context, which has been used across the seminal word vector-learning algorithms.

\section{Representation and Co-Occurrence} \label{sec:RepCo}
Harmonic distributional structures have long been observed, but applications of them to NLP systems have largely not emerged. We can juxtapose this lack of application to the transformative impact on NLP by representation learning's \textit{embeddings}, or, \textit{word vectors}. These allow modern DL systems to approximate the meanings of tokens. Since Latent Semantic Analysis (LSA) was introduced~\cite{dumais1988using}, vector representations of tokens have been used to predict and retrieve synonyms and analogies \cite{mikolov2013efficient,mikolov2013distributed,pennington2014glove}. The fact that word vectors exhibit linear semantic relationships between tokens, i.e., predict analogies, is heralded as a success in their capture of meaning, but exists without solid understanding of how these meanings are captured. LSA has influenced theories about human cognition~\cite{landauer1997solution}
and been used to measure association of concepts during free recall \cite{howard2002does,zaromb2006temporal}.
Word vectors are limited in representing polysemous words. However, as demonstrated in \cite{arora2018linear}, polysemous words lie in a superposition of their senses within a linear semantic space, and one can approximately recover underlying sense vectors \cite{arora2018linear}.

\subsection{Modeling Co-Occurrence}
The statistical dynamics of co-occurrence strongly depend on the hyper-parameter $m$, whose effect can be seen from a low-complexity model. Specifically, one can crudely sample from an empirical, harmonic-frequency distribution to retain some realistic structure. To compute a model $\hat{F}^m$ for a document collection, $\mathcal{D}$, a token $t$ samples $f_t$ windows of $\pm m$ other tokens $s$ that are also distributed by $f$. This makes the sampling proportional to frequency ratios with $t$: $\hat{F}^m_{t,s}~=~C_tf_s/f_t$. To physicalize the model, one need only assert: $\sum_{s\in V}\hat{F}^m_{t,s} = 2mf_t$ and solve for the constant of proportionality, $2mf_t^2/M$, allowing for a closed-form specification:
\begin{equation}
\hat{F}^m_{t,s} = \frac{2mf_tf_s}{M}
\label{eq:X-solution}
\end{equation}
We refer to \textbf{Eq.~\ref{eq:X-solution}} as the \textit{independent frequencies model} (IFM), which forms a dense co-occurrence matrix that is computable from \textit{any} set of unigram frequencies. To view this model, we present \textbf{Fig.~\ref{fig:noise-words}}, which exhibits the IFM against co-occurrences of the word `they' in the Georgetown University Multilayer (GUM) Corpus~\cite{zeldes2017}.

\subsection{Co-Occurrence Factorizations}
There is a deep connection between word representation algorithms and the factorization of token co-occurrence matrices. This connection is perhaps most transparent for the GloVe algorithm~\cite{pennington2014glove}, whose loss function is \textit{defined} to factor the positive values of the $\log$-co-occurrence matrix, and is minimized under frequency-dependent weights, $W$, to produce word vectors $\vec{u}_t,\vec{v}_s$ and bias parameters $a_t,b_s$ that predict the values of $F^m$:
\begin{equation}
\sum_{t,s\in\mathcal{D}}W_{t,s}\left(\vec{u}_t\vec{v}_s^T + a_t + b_s - \log{F^m_{t,s}}\right)^2
\label{eq:GloVe-loss}
\end{equation}

Under GloVe's loss function (Eq.~\ref{eq:GloVe-loss}), a perfect model's point of convergence would have zero-valued squared terms~\cite{kenyon-dean2020}: 
\begin{equation}
    \log{F^m_{t,s}} = \vec{u}_t\vec{v}_s^T + a_t + b_s
    \label{eq:GloVe-factorization}
\end{equation}
Observing this point of convergence, \cite{kenyon-dean2020} remark upon the variation exhibited by GloVe's vector products and bias terms, but provide little insight into \textit{how} word vectors interact via inner products to produce PMI-like values. We investigate these details and discover critical, mechanical insights that will be used to produce a bias-probing methodology. 

\subsubsection{Clamped GloVe}
Separating the effects of bias terms and vector products is essential for understanding GloVe's connection to other models, and can be achieved by introducing a `clamping' hyper-parameter, $\kappa\in{\{0,1\}}$, to turn on/off the bias terms. Multiplying this Boolean factor into the bias terms, GloVe's general factorization is:
\begin{equation}
    \log{F^m_{t,s}} = \vec{u}_t\vec{v}_s^T + \kappa(a_t + b_s)
    \label{eq:GloVe-factorization-clamped}
\end{equation}
So, suppose GloVe is clamped ($\kappa = 0$) and that its data follow the IFM (Eq.~\ref{eq:X-solution}). In this case, \textit{vector differences}, e.g., between $\vec{u}_t$ and $\vec{u}_s$, act on \textit{every} other token $w\in \mathcal{V}$'s $v$-vector as a constant: $\left(\vec{u}_t - \vec{u}_s\right)~\cdot~\vec{v}_{w}^T~=~\log(f_s/f_t)$. This then indicates that pairs of vectors with the same frequency ratio: $\frac{f_{s_x}}{f_{t_x}}~=~\frac{f_{s_y}}{f_{t_y}}$ 
have representations which operate semantically equivalently, under the GloVe model. We now emphasize the importance of this \textit{frequency-ratios property} in describing model mechanics \textit{across all} classical word-vector models. 

\subsubsection{The Frequency-Ratios Property}
As it will be regularly discussed throughout the remainder of this document we formally define the \textit{frequency-ratios property} for \textit{any} classical word-vector representation, below.

\vspace{5pt}\noindent\textbf{Definition}: Given two words from a vocabulary $t, s\in \mathcal{V}$ and any set of classical, IFM-trained word vectors: $U,V\in R^{|\mathcal{V}|\times k}$ ($k\leq |\mathcal{V}|$), the \textit{frequency-ratios property} exists when the action of \textit{vector differences}, e.g., between $U$-vectors on \textit{any} other token $w\in \mathcal{V}$'s $V$-vector is equal to the $\log$-frequency ratio of $t$ and $s$, regardless of $w$'s choice: 
\begin{equation}
\left(\vec{u}_t - \vec{u}_s\right) \cdot \vec{v}_{w}^T =
\log\frac{f_s}{f_t}
\label{eq:FRP}
\end{equation}

\vspace{5pt}\noindent We'll use the frequency-ratios property to efficiently measure the semantic bias in data that representations \textit{would} learn. To get there, we will ultimately ask: does the linear-semantic analogy property (completing analogies by addition/subtraction) relate to a relationship of comparable unigram frequency ratios (and products)? For example, this asks if ``man~is~to~king~as~woman~is~to~queen'' is described in data by: $f_\text{king}/f_\text{man}\approx~f_\text{queen}/f_\text{woman}$.

\subsubsection{Un-Clamped GloVe}
While a clamped model is technically less complex (having fewer predictive parameters), GloVe is often defined without clamping. In this case ($\kappa = 1$), the connections between vector differences and model parameters become less clear. Current conjecture inclines bias parameters will converge to $\log$-unigram-frequency values, leaving the vector products to model the point-wise mutual information (PMI) between tokens~\cite{kenyon-dean2020}. Provided GloVe's inner products model the PMI, training GloVe on the IFM (Eq.~\ref{eq:X-solution}) should force all vector products to zero: $\vec{u}_t\vec{v}_s^T\hspace{-2.5pt}~=~\hspace{-2.5pt}0$. From this view, \textit{more-independent word co-occurrences should produce less informative vector products, i.e., poorer GloVe models.} While the evidence for the bias-parameters' dependence on frequency is compelling, we note that in \cite{kenyon-dean2020}'s comparison of bias terms with unigram frequency exhibited a \textit{super-linear} trend, which from the logarithmic scale of presentation allows rough approximation by a power-law. Denoting a model exponent by $\gamma>1$, one can estimate the un-clamped bias parameters' behavior as $e^{a_t},e^{b_s} \propto (f^m_t)^\gamma, (f^m_s)^\gamma$. With an IFM defined by $F^m$-marginal frequencies, GloVe's $\gamma$-scaled PMIs exhibit a frequency-ratios property: $\left(\vec{u}_t-\vec{u}_s\right)~\cdot~\vec{v}_{w}^T=~(\gamma-1)\log(f^m_s/f^m_t)$. So GloVe's optimization away from \textit{true} PMI avoids inner-product singularities under the IFM, ensuring the frequency-ratios property's presence.

\begin{figure}
  \centering
  \includegraphics[width=0.99\linewidth]{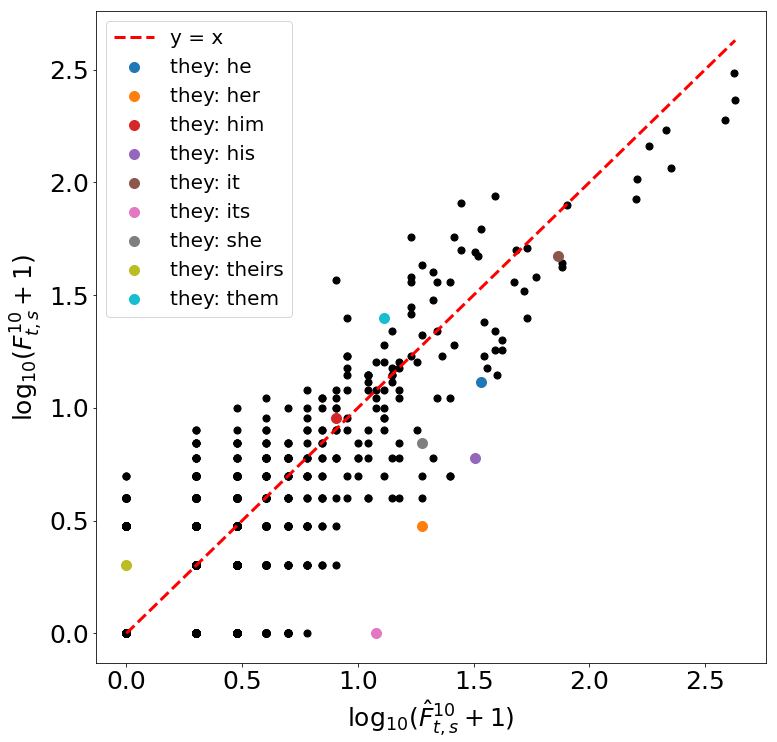}
  \caption{Comparison of the IFM and empirical co-occurrences for the word ``they'' within the GUM corpus. Statistical dependencies between words are the distances between points and the red dashed line. Unity is added to all points to clarify non-occurrent pairs.}
  \label{fig:noise-words}
\end{figure}

\subsubsection{Word2Vec Softmax}
Prior to GloVe's development, the Word2Vec algorithm first emerged as a seminal advancement for word representation. While Word2Vec is perhaps most commonly applied under the skip-gram with negative sampling (SGNS) objective~\cite{mikolov2013}, negative sampling objectives were originally developed to approximate more computationally complex softmax objectives~\cite{mikolov2013efficient}. Here, we investigate the effects of the IFM's co-occurrences on both objectives.

While it has been known for some time that the SGNS Word2Vec objective factorizes a shifted PMI matrix~\cite{levy2014neural}, the implicit matrix factorization behind Word2Vec's softmax objective to-date has not been derived. While this could be due to the softmax objective's mathematical complexity or the perceived lack of a factorization's utility (given unfactorized softmax's computational complexity), we show that neither is truly an obstacle and now derive the softmax factorization. While providing insight into \textit{Word2Vec as an LM}, this presents an optimization strategy that makes the softmax objective much more computationally feasible, opening new potential for large scale applications, which we leave to future work.

\vspace{5pt}\noindent\textbf{Theorem}: Under the $\log$-softmax objective: 
\begin{equation}
\mathcal{L}_{soft} = -\sum_{t\in V}\sum_{s\in V}F^m_{t,s}\log\varphi(\vec{u}_t\vec{v}_s),
\label{eq:soft-objective}
\end{equation}
the Word2Vec algorithm implicitly converges towards a matrix factorization for all non-zero co-occurrences of the form:
\begin{equation}
\vec{u}_t\vec{v}_s^T =  \log \frac{F^m_{t,s}}{f^m_t},    
\label{eq:soft-factorization}
\end{equation}
which is equal to the $\log$-conditional probability matrix of the co-occurrence model.

The proof of this theorem is provided in \textbf{Appendix~\ref{sec:softmax-limit}}, which is the first known---to the best of our knowledge---proof of the softmax objective's factorization. This facorization produces a true LM, while Word2Vec's SGNS objective and GloVe do not. Historically, the softmax objective hasn't been utilized for pre-training applications due to computational complexity, and because SGNS has been seen as a partial approximation of softmax. However, our softmax solution provides a low-complexity strategy for pre-training---more powerful---\textit{LM}-representations efficiently, via aggregated co-occurrences, just like GloVe's regression-based loss, e.g., replacing the rightmost term of \textbf{Eq.~\ref{eq:GloVe-loss}} with \textbf{Eq.~\ref{eq:soft-factorization}}. This could have far reaching consequences, but devising new pre-training techniques was not the explicit intention of this proof. For us, the factorization elucidates the existence of the frequency-ratios property for what is arguably the most fundamental/influential classical word vector algorithm as a corollary (proof, \textbf{Appendix~\ref{sec:softmax-ratios}}).

\vspace{5pt}\noindent\textbf{Corollary}: When trained on the IFM, Word2Vec's softmax objective, $\mathcal{L}_{soft}$, exhibits the frequency-ratios property asymmetrically for differences of $V$-vectors acting on $U$ (only).

\vspace{5pt} Similar to \textit{un-clamped} GloVe, the softmax skip-gram objective for Word2Vec only supports the frequency-ratios property on one side. The other side of its parameters could be responsible for maintaining the softmax's normalization and/or contrast. This could possibly explain why one of the $U$ vs. $V$ matrices' parameters have traditionally been preferentially retained, i.e., since only the $V$-vector differences are guarenteed to exhibit the frequency-ratios property when acting on $U$. However, we note that both $U$ and $V$ are intrinsically intertwined as two complimentary parts of the factorization. This also indicates that both $U$ and $V$ matrices should probably be retained for later use, and perhaps only ever combined by concatenation, since distribution in this form would allow other researchers apply full models. For example, this would allow for \textit{vectors} trained by \textbf{Eq.~\ref{eq:soft-factorization}} to be used \textit{as a low-compute LM}.

\subsubsection{Word2Vec-SGNS}
We ask if the SGNS-Word2Vec objective also exhibits any frequency-ratios property. Here, we find asymmetric support again, and which is strikingly similar to that of un-clamped GloVe:

\vspace{5pt}\noindent\textbf{Theorem}: The Word2Vec SGNS objective trains vectors which exhibit a frequency-ratios property scaled by one minus its sampling parameter: $\vec{u}_w(\vec{v}_{t}-\vec{v}_{s})^T=(1-\alpha)\log(f^m_s/f^m_t)$.

\vspace{5pt}\noindent This theorem (proof, \textbf{Appendix~\ref{sec:SGNS-ratios}}) shouldn't be too surprising, since SGNS \textit{also} factors a PMI-like matrix~\cite{kenyon-dean2020}. What is perhaps most surprising about this result is that SGNS' frequency-ratios property emerges directly from hyper-paramaterization via $\alpha>0$, which tempers the negative-sampling rate as a power-law scaling of frequency. While $\alpha$ is generally presented with limited theoretical justification, its intent is accelerated learning, and its effect is biased (high-entropy) sampling during learning. Reflecting on this, it seems possible that un-clamping Glove induces $\gamma$ in lieu of receiving a biased sample of contrastive information via $\alpha$, as is done with SGNS. We likewise note that the piece-wise construction of $W$ in GloVe's formulation complicates analysis, which could explain $\gamma$'s limited presence over only the largest frequencies~\cite{kenyon-dean2020}.

Considering how the frequency-ratios property appears ubiquitously across the diversity of classical word vector models \textit{under the IFM}, we will examine the degree to which independence pervades co-occurrence models, below. However, with the frequency-ratios property in the focus, we now exhibit its immediate capacity to profile the semantic biases present in data.

\section{Probing Data for Semantic Biases}
The experiments described here draw from several publicly available data sets and intend to exhibit how analogies and token frequencies interact. Token-frequency distributions are taken from two corpora denoted by $\mathcal{G}$ and $\mathcal{W}$, corresponding to Google Books' most recent $N$-grams release~\cite{GoogleNgrams} and a controlled collection of Wikipedia articles, described in detail below. Our interest with analogies is not in their prediction, and rather in developing a bias-probing methodology for evaluating \textit{data}. So while a number of analogical test sets exist---including from the well-studied MSR collection~\cite{mikolov2013efficient,mikolov2013distributed}---we utilize the Bigger Analogy Test Set (BATS) for its size, organization, and diversity, providing a total of roughly $10^5$ analogical comparisons across categories~\cite{gladkova2016}. 

Critically, BATS contains analogical comparisons for multiple encyclopedic groups. While analogical prediction experiments often perform well at the country-capital relationship, the more acute geographic category comparing UK counties and cities appears more challenging~\cite{gladkova2016}. We ask if this lower performance is due to poor representation in source data, i.e., if the relative abundance of language which discusses UK cities and counties \textit{is low} in the data used to train word vectors that have been studied in the past. To examine this question, we will study the extent to which an intentionally-biased sample exhibits support for the UK city-county analogies.

\subsection{Bias Measurement via Analogies}
Stepping back, there should be no surprise if analogies can be used to directly measure bias in data. The WEAT test for measuring bias in word vectors~\cite{caliskan2017} is based on four same-sized sets of words, which are referred to as target, e.g., gender-related words; and attribute, e.g., role-related words. Sampling one word from each of these sets essentially forms an analogy (even if non-sensical or offensive), and the WEAT formula measures bias via similarity statistics averaged across all comparisons. Furthermore, more recent methods for controlling bias in modern, self-attending systems retain this formulation~\cite{karve2019}, generalizing WEAT to four potentially-different-sized word sets, but again, with two for target words and two for attributes that can be used to constitute the dyads of hypothetical analogies.

To measure bias directly in data using analogies, denote each of the dyads, e.g., $(\text{man}, \text{woman})$, within a given analogy, $x:y$, as: $x = (t,s)$ and $y = (\tilde{t},\tilde{s})$. On any given pair of dyads, we introduce the absolute difference of $\log$-frequency ratios as a measurement of the \textit{dissonance}, $\Delta$, expressed towards the dyads, given a corpus, $\mathcal{D}$:
\begin{equation}
    \Delta\left(x,y\mid\mathcal{D}\right) = 
    \left|\log\frac{f_tf_ {\tilde{s}}}{f_sf_{\tilde{t}}}\right|/
    \text{max}_{l\in\mathcal{V}}\{\log f_l\}
    \label{eq:dissonance}
\end{equation}
This quantity is entropicly `normalized' by the largest value that the absolute difference could possibly take, occurring when a dyad of ratio $1$ is compared to one with least-most frequent words. This places $\Delta$ in $[0,1]$ and makes it possible to compare dissonance \textit{between} corpora, i.e., the expressions of bias that data exhibit. 

Musically, $\Delta$, measures the degree to which the dyads are consonantly/dissonantly equivalent, i.e., whether the dyads play the same `chord' (regardless of pitch). This is because $\Delta$ can be considered in terms of physical waves, i.e., modeling a document as a superposition of unit-amplitude square waves, whose peaks approximate the positions of each type's occurrences. Since physical waves of constant amplitude and velocity will have powers proportional to the squares of their frequencies, each token-frequency ratio becomes equivalent to the square root of two waves' power ratio. Thus when un-normalized, the dissonance can be understood by its units of \textit{decibels}, which more broadly informs us that $\Delta$ measures an absolute difference in decibels expressed by each dyad, or, \textit{the difference in loudness between the dyads' overtones}.

\begin{figure*}[t!]
  \centering
  \includegraphics[width=0.99\textwidth]{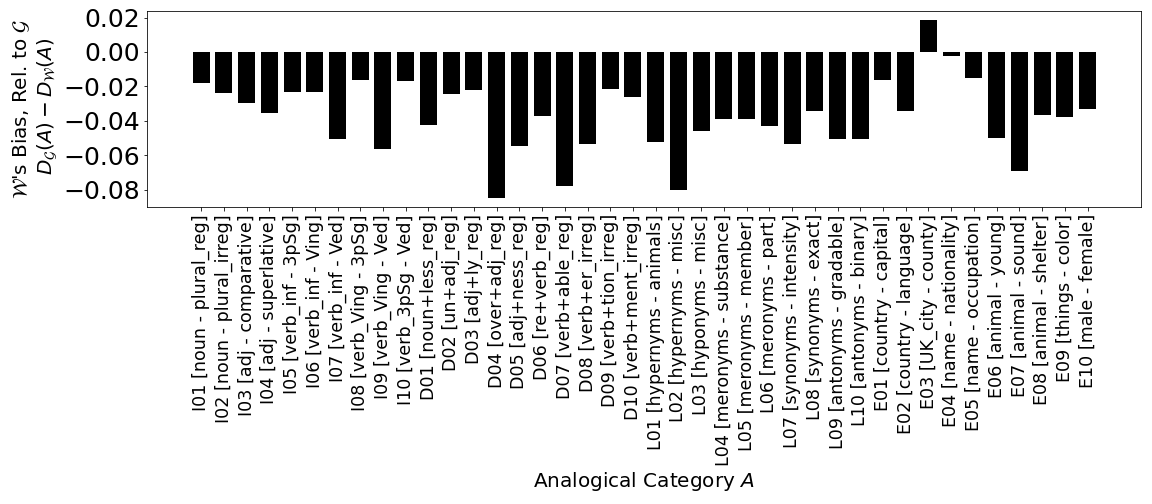}
  \caption{Comparison of the dissonance towards the different BATS analogy categories for the Google Book corpus, $\mathcal{G}$, and a much smaller corpus of Wikipedia articles that connect to pages discussing the UK cities. Positive bars indicate categories towards which $\mathcal{W}$ is more biased, i.e., which contain analogies that $\mathcal{W}$ supports more.}
  \label{fig:consonance}
\end{figure*}

\subsection{Analogical-Bias Probing Experiment}
While it is customary to train word vectors on Wikipedia articles, we hypothesize that historical samples have had relatively few descriptive passages relating the UK cities to their counties, as compared to national capitols with their countries. The latter are likely more-broadly discussed on Wikipedia, and we hypothesize the former would be if their relative associations were more adequately represented in data. To ensure this, we composed our sample of Wikipedia articles, $\mathcal{W}$, from the collection of \textit{all pages leading to and from any UK city's Wikipedia page,} where UK cities are strictly defined according to those listed on Wikipedia in its presentation of the UK's 69 officially-designated cities (as of 2021). This resulted in a Wikipedia corpus of roughly $200,000$ articles that is linked to the subjects of UK cities and counties.\footnote{Accessed 10/31/21: \url{en.wikipedia.org/wiki/List_of_cities_in_the_United_Kingdom}} By comparison of frequencies, $\mathcal{W}$ is about one-thousand times smaller than $\mathcal{G}$.

In application, low-dissonance values indicate which analogies are supported by corpus frequency ratios. To determine the \textit{overall} support a corpus has for a \textit{set} of analogies, $A$, an average, $D_\mathcal{D}(A)$, of $\Delta$-dissonance values can be computed. Methodologically, we weight averages by the corpus frequencies of the tokens within each analogy's dyads. In \textbf{Fig.~\ref{fig:consonance}}, we compute the difference of averages between $\mathcal{G}$ and $\mathcal{W}$: $D_\mathcal{G}(A) - D_\mathcal{W}(A)$ for the different analogical categories of BATs, i.e., so that positive bars indicate where $\mathcal{W}$ supports a category's analogies more than $\mathcal{G}$. Average values on their own (no differences) for this experiment can likewise be observed within \textbf{Tab.~\ref{tab:dissonance}} in the Appendices. Either view exhibits how the \textit{only} category for which $\mathcal{W}$ exhibits less dissonance (more bias) than $\mathcal{G}$ is the UK city-county category, and furthermore, that this bias is clustered amongst related categories, e.g., name-nationality, which are elevated. We view these results as quite sensible for a bias metric, and indicating a promising pathway towards developing low-cost and -compute bias probes for data. This will feature centrally in our final discussion, after investigating the IFM's relevance to real-world corpora, both for its central role in elucidating the analogical bias probe methodology, and the paths it lays toward future discoveries.

\section{Co-occurrence and Independence}
Our study of the IFM and measurement of bias in data with analogical test sets raises an important question: \textit{how relevant is the IFM to real-world data?} While it's not possible to objectively state if co-occurrences are independent or not, empirical systems do express independence on a spectrum. Determining the prevalence of independence in co-occurrence statistics requires control over the context model, i.e., $m$ affects independence. To compare co-occurrence frequencies between the data and those sampled independently from marginal distributions, one can compare to independence, as measured by the IFM (\textbf{Fig.~\ref{fig:noise-words}}). In the figure, the extent to which the empirical frequencies are equal to the IFM's can be quantified by how close points fall to the line $y=x$. Intuitively, this expresses how independent the empirical co-occurrences are when $m=10$, and grounds the subject (independence) that we wish to study at larger scales of data.

\subsection{Quantifying Independence}
To measure statistical dependencies one can take the PMI's expectation over its joint probabilities and compute the co-occurrence mutual information (MI). MI measures how dependent the statistics of a joint distribution are. When MI is normalized by the joint information its values fall in $[0,1]$ and define the information quality ratio (IQR): 
$\mathcal{I}_{k,m} = -{\sum_{t,s\in V}P^m_{t,s}\log\frac{P^m_{t,s}}{p^m_tp^m_s}}/{\sum_{t,s\in V}P^m_{t,v}\log P^m_{i,j}}$.
Each of $P^m$ and $p^m$ are probabilistic forms of $F^m$ and $f^m$ (divided by $M^m_F$), and we will used $k$ to record the number of documents in a given sample. Intuitively, $\mathcal{I}_{k,m}$ describes how close to independent co-occurrences are, and  $\mathcal{I}_{k,m} \rightarrow 0$ indicates co-occurrences becoming more independent. 

Linguistic dependencies are reported to exhibit power-law relationships between dependence length-$d$ (the number of other words up through the dependence) and frequency~\cite{chen2019}. Through preliminary analysis of the GUM corpus', we find $\mathcal{I}_{k,m}$ values---statistical dependencies---\textit{also} appear to decay as a power law function of the window size, $m$ (\textbf{Appendix~\ref{sec:modeling-dependencies}}). Critically, we observe that the sentence-length distribution sets a bound on the background of contrastive information, saturating with increasing values of $m$ (\textbf{Fig.~\ref{fig:GUM}}). However, for even the smallest-$k$ corpora and \textit{any} values of $m$, $\mathcal{I}_{k,m}$ appears less than $0.5$, suggesting in some sense that \textit{co-occurrences are more independent than dependent}. However, this view is reductive and leaves some critical questions, which we begin to address below.

\noindent\textbf{Does the IQR measure linguistic dependencies?} Beyond observing $\mathcal{I}_{k,m}<0.5$, we evaluate $\mathcal{I}_{k,m}$ more broadly in \textbf{Appendix~\ref{sec:independence-appendix}}, where a full profile of $\mathcal{I}_{k,m}$-values is provided for GUM. We likewise model the IQR as a power-law function of the context-window size, $m$, to elucidate if and how linguistic dependencies contribute to the statistical dynamics of co-occurrence dependence, as measured by $\mathcal{I}_{k,m}$. Seemingly, $\mathcal{I}_{k,m}$ can be modeled strongly, when properly modulated by the sentence length distribution. This means sentence tokenization plays a central role in defining co-occurrence statistics. Likewise, lower-quality sentence tokenizations seem to result in more-complex distributions, optimization challenges, and noisy $\mathcal{I}_{k,m}$-profiles. Perhaps most surprisingly, the power-law which models $\mathcal{I}_{k,m}$'s \textit{statistical dependencies} appears to exhibit a scaling exponent, $\nu$, which parametrically fits the density of \textit{linguistic dependencies} annotated in GUM (\textbf{Fig.~\ref{fig:GUM}}, inset). Further experimentation on different, parsed corpora is clearly required to determine if this model and relationship are robust.

\noindent\textbf{Does the IQR have a lower bound?} Determining this requires measuring and fitting $\mathcal{I}_{k,m}$ for larger data sets. Combinatorality imposes significant computational challenges for large values of $m$, so lower-$m$ values (a smaller window) were used to measure the IQR for larger values of $k$ (with more data). As there is no way to measure $\mathcal{I}_{k,m}$ for arbitrary corpus sizes, limiting arguments are ultimately required (\textbf{Appendix.~\ref{sec:IQR-bound}}). We find that the $k$-limiting dynamics of $\mathcal{I}_{k,m}$-values appear non-zero and convergent with bounds that can be solved (\textbf{Eq.~\ref{eq:IQR-bound}}) and computed (\textbf{Fig.~\ref{fig:WikiText}}). 

\section{Discussion and Conclusions}
The gravity of the IQR's lower bound should not be understated: even a countably-infinite collection of documents will retain a definite portion of dependent statistical information in its co-occurrences. In some sense, this assures the statistical need for large corpora to `chip away' at the underlying statistical dependencies recorded in linguistic data. However, while convergence is rapid at first, it slows considerably for larger corpora, indicating ever-diminishing returns from bigger data. From our bias-probing experiments, we exhibit how more data isn't necessarily more representative (\textbf{Fig.~\ref{fig:consonance}}). Thus, we ask if the IQR's limiting behavior is a process of document structure washing out in favor of more-local relationships. If so, we might then interpret \textbf{Fig.~\ref{fig:consonance}} as exhibiting a corpus whose co-occurrences have been unusually constrained for its document distribution, providing another interpretation of semantic bias. 

Seemingly, statistical dependencies are sparse in sentences, and $m$-word sliding-window context models can't separate these from independent variation while absorbing co-occurrences. This interpretation can be intuitively stated by modifying Firth's famous quote (hence the paper's title): \textit{You shall know a word by the company it keeps, and what else lies in the vicinity.} However, we know less if it is us who know words by this truism, as much as it is AIs who \textit{know} how to use words by it.

Arriving at this point has entailed the development of novel techniques for probing unstructured, linguistic data for semantic biases using data sets of analogies. On their own, these results appear positive, and exhibit their own methodological value. However, they likewise emerged from another discovery, of the universal, frequency-ratios property for word vectors. The substantiation of that property required deriving a limiting factorization for the \textit{original} Word2Vec objective, whose apparent natural formulation as \textit{the} contextualizing LM produced by the co-occurrence conditional-probably-matrix underpins its importance. This production of a closed-form solution to Word2Vec could perhaps produce the biggest impacts of this work by providing rich new pathways towards efficiently deriving representation statistics. To this end, we highlight the IFM's derivation as another core outcome of our work, both for its central roles in analysis, and it's potential to warm-start embedding layers efficiently via unigram statistics.


\bibliography{custom}
\bibliographystyle{acl_natbib}

\appendix

\section{Word2Vec's Softmax Factorization}
\label{sec:softmax-limit}

\vspace{5pt}\noindent\textbf{Theorem}: Under the $\log$-softmax objective: 
\begin{equation}
\mathcal{L}_{soft} = -\sum_{t\in V}\sum_{s\in V}F^m_{t,s}\log\varphi(\vec{u}_t\vec{v}_s),
\label{eq:soft-objective-appendix}
\end{equation}
the Word2Vec algorithm implicitly converges towards a matrix factorization for all non-zero co-occurrences of the form:
\begin{equation}
\vec{u}_t\vec{v}_s^T =  \log \frac{F^m_{t,s}}{f^m_t},    
\label{eq:soft-fact-appendix}
\end{equation}
which is equal to the $\log$-conditional probability matrix of the co-occurrence model.

\vspace{5pt}\noindent \textbf{Proof}: The softmax function is computed by row: $\varphi(\vec{u}_t\vec{v}_s^T) = e^{\vec{u}_t\vec{v}_s^T}/\sum_{l\in V}e^{\vec{u}_t\vec{v}_l^T}$. To solve for $\vec{u}_t\vec{v}_s^T$, we must determine all components of $\mathcal{L}_{soft}$'s gradient which depend on $\vec{u}_t\vec{v}_s^T$, and which arise from different portions of $\mathcal{L}_{soft}$'s Jacobian. This includes the positive, differential portion from the softmax's numerator: $-F^m_{t,s}\left(1 - \varphi\left(\vec{u}_t\vec{v}_s^T\right)\right)$ as well as the negative, differential portion emerging from the softmax denominators: $\sum_{l\in V\setminus\{s\}} F^m_{t,l}\varphi\left(\vec{u}_t\vec{v}_s^T\right)$,
which sums over all $l\neq s$, since softmax's derivative is vector valued.

By combining the negative and positive portions, the partial derivative of $\mathcal{L}_{soft}$ with respect to $\vec{u}_t\vec{v}_s^T$ is a sum which ranges over the entire vocabulary:
$$
\frac{\partial\mathcal{L}_{soft}}{\partial \left(\vec{u}_t\vec{v}_s^T\right)}\left(\vec{u}_t\vec{v}_s^T\right) = 
-F^m_{t,s} + \sum_{l\in V} F^m_{t,l}\varphi\left(\vec{u}_t\vec{v}_s^T\right)
$$
When set equal to zero, the sum is easily solved:
\begin{equation}
\varphi\left(\vec{u}_t\vec{v}_s^T\right) = 
\frac{F^m_{t,s}}{\sum_{l\in V} F^m_{t,l}} = 
\frac{F^m_{t,s}}{f^m_t}
\label{eq:softmax-factorization}    
\end{equation}
where the co-occurrence $m$-window `inflation' defines the unigram statistics as: $f^m_t = \sum_{l\in V} F^m_{t,l}$ by the $t^\text{th}$ marginal sum, i.e., pushing the factorization towards the $\log$-conditional probability matrix

\textbf{Eq.~\ref{eq:softmax-factorization}} almost provides the main result, but only factorizes the softmax's application. Due to normalization, there will necessarily be error from the $\log$-conditional probability matrix, 
which we handle by defining some $\beta_t$ close to $1$ in $(0,1)$ for each $t\in V$. Selecting these values can used to produce an ansatz solution, which can be used to understand the limiting matrix being factorized, and hence algebraically solve for arbitrarily-well optimized softmax representations. First, define the ansatz's positive-occurring elements by:
$$
\vec{u}_t\vec{v}_s^T= \log \beta_t\frac{F^m_{t,s}}{f^m_t}
$$
Then define $t$'s $n_t$ negative-occurring elements by:
$$
\vec{u}_t\vec{v}_l^T = \log\frac{1 - \beta_t}{n_t}
$$
Under this initialization, the error for the positive-occurrence pairs, $\varepsilon_{t,s}$, is determined as:
$$
\varepsilon_{t,s} = 
\frac{F^m_{t,s}}{f^m_t} - \varphi\left(\vec{u}_t\vec{v}_s^T\right) =
\frac{F^m_{t,s}}{f^m_t}(1-\beta_t)
$$
Likewise, we can also now easily observe the error for the non-occurrent pairs:
$$
\varepsilon_{t,l} = 
-\varphi\left(\vec{u}_t\vec{v}_l^T\right) =
\frac{1-\beta_t}{k_t}
$$
Critically, these errors diminish as $\beta_t\rightarrow 1$. Furthermore, driving $\beta_t\rightarrow 1$ reduces the negative log likelihood as it pushes the true co-occurrent factorized values towards the claimed limiting solution. This also indicates that the softmax model likely has no exact algebraic solution for its factorization. Specifically, while positive-occurring entries converge toward:
$$
\lim_{\beta_t\rightarrow 1} \vec{u}_t\vec{v}_s^T = \log \frac{F^m_{t,s}}{f^m_t}
$$
the non-occurring pairs in the factorized matrix have values which become ever more negative:
$$
\lim_{\beta_t\rightarrow 1} \vec{u}_t\vec{v}_l^T = -\infty
$$
This is generally the case for GloVe and Word2Vec's SGNS objective, too, as neither is defined on negative-occurring values and would require a similar, negative-diverging ansatz for an algebraic solution to their factorizations. This concludes the main proof, and now allows for investigation of how unigram-frequency ratios interact with vector differences. $\blacksquare$

\section{Softmax Frequency Ratios}
\label{sec:softmax-ratios}
\vspace{5pt}\noindent\textbf{Corollary}: When trained on the IFM, Word2Vec's softmax objective, $\mathcal{L}_{soft}$, exhibits the frequency-ratios property asymmetrically for differences of $V$-vectors acting on $U$ (only).

\vspace{5pt}\noindent\textbf{Proof}: Substituting the IFM into the solved softmax-Word2Vec factorization (\textbf{Eq.~\ref{eq:soft-fact-appendix}}), we find:
$$
\vec{u}_t(\vec{v}_s - \vec{v}_{\tilde s})^T= \log\frac{f_s}{f_{\tilde s}}
$$
which is precisely the frequency-ratios property. However, when we apply this analysis symmetrically we find  something different: 
$$
(\vec{u}_t - \vec{u}_{\tilde t})\vec{v}_s^T = 
\log \frac{\beta_tf_tf^m_{\tilde t}}{\beta_{\tilde t}f_{\tilde t}f^m_t},
$$
which depends on co-occurrence `inflation', as well as the ansatz's choice of $\beta_t$ values. This is yet another, different form of the frequency-ratios property, where if the $\beta_t$ values are chosen proportional to their respectively-\textit{inflated} unigram frequencies: $\beta_t = f^m_t/M^m_F$, the exact frequency-ratios property is recovered to a full symmetry. $\blacksquare$

\section{SGNS Frequency Ratios}
\label{sec:SGNS-ratios}

\vspace{5pt}\noindent\textbf{Theorem}: The Word2Vec SGNS objective trains vectors which exhibit a frequency-ratios property scaled by one minus its sampling parameter: $\vec{u}_w(\vec{v}_{t}-\vec{v}_{s})^T=(1-\alpha)\log(f^m_s/f^m_t)$.

\vspace{5pt}\noindent\textbf{Proof}: In \cite{kenyon-dean2020}'s work, the noise distribution was assumed different from convention, which utilizes a hyper-parameter, $\alpha\in R$, commonly set to $\alpha = 3/4$. Its general effect will modify the PMI-convergence points into:
$$
\vec{u}_t\vec{v}_s^T = 
-\log\left[\frac{F^m_{t,s}}{f^m_t}\frac{M_f^{m,\alpha}}{\left(f^m_s\right)^{\alpha}}\right] + \log k.
$$
where $M_f^{m,\alpha}$ is the corresponding normalization constant for an $\alpha$-power, $m$-inflated unigram frequency distribution: $M_f^{m,\alpha} = \sum_{s\in V}\left(f^m_s\right)^\alpha$. 

Observing the $V$-vector-difference action on $U$ and 
When one defines an IFM by the inflated unigram statistics and
substitutes the corresponding $\hat{F}^m$in for $F^m$ another frequency-ratios property emerges, but this time with effect scaled by $1 - \alpha$:
$$
\vec{u}_w(\vec{v}_t - \vec{v}_s)^T =
(1 - \alpha)\log\frac{f^m_s}{f^m_t}
$$
So when one sets $\alpha\neq 1$, the frequency-ratios property appears again, for Word2Vec's $V$-vectors on the $U$ matrix. Note that the only value of $\alpha$ for which this property doesn't exist ($\alpha = 1$) is generally not utilized in applications, with most discussion usually asserting that $\alpha = 1$ produces less adept models.

For SGNS, a frequency-ratios property is only clearly entailed for $U$-vector differences on the $V$ matrix. However, it appears that the frequency-ratios property for $U$-vector differences on $V$ should be absent, as SGNS's negative information/normalization is noisily rigid, based entirely on independent sampling at a fixed rate of $k$-to-$1$:
$$
(\vec{u}_t - \vec{u}_s)\vec{v}_w^T =
-\log\frac{\hat F^m_{t,w}f^m_s}{\hat{F}^m_{s,w}f^m_t} = 0
$$
A fuller analysis of the frequency-ratios property for the SGNS objective (as well as for softmax) would ultimately benefit from limiting analysis of the gradient descent process. While this is partly considered for softmax in the context of an ansatz solution, further discussions of limiting effects and optimization is left for future work.  $\blacksquare$

\newpage 
\section{Bias Probing Experiment}
\label{sec:analogical-dissonance}

\begin{table}[b!]
    \centering
    \begin{tabular}{|l|c|c|}
        \hline
        \multicolumn{1}{|c|}{\textbf{BATS Category}} & \multicolumn{2}{|c|}{\textbf{Dissonance} ($D$)}\\\hline
        \textbf{Inflection} & $\mathcal{G}$ & $\mathcal{W}$\\\hline
        I01: noun-plural\_reg & \textbf{0.035} & 0.053 \\
        I02: noun-plural\_irreg & \textbf{0.044} & 0.068 \\
        I03: adj-comparative & \textbf{0.038} & 0.068 \\
        I04: adj-superlative & \textbf{0.032} & 0.068 \\
        I05: verb\_inf-3pSg & \textbf{0.033} & 0.057 \\
        I06: verb\_inf-Ving & \textbf{0.03} & 0.054 \\
        I07: verb\_inf-Ved & \textbf{0.025} & 0.076 \\
        I08: verb\_Ving-3pSg & \textbf{0.048} & 0.064 \\
        I09: verb\_Ving-Ved & \textbf{0.036} & 0.092 \\
        I10: verb\_3pSg-Ved & \textbf{0.04} & 0.057 \\\hline
        \textbf{Derivation} & $\mathcal{G}$ & $\mathcal{W}$\\\hline
        D01: noun+less\_reg & \textbf{0.076} & 0.118 \\
        D02: un+adj\_reg & \textbf{0.048} & 0.072 \\
        D03: adj+ly\_reg & \textbf{0.045} & 0.067 \\
        D04: over+adj\_reg & \textbf{0.066} & 0.151 \\
        D05: adj+ness\_reg & \textbf{0.074} & 0.129 \\
        D06: re+verb\_reg & \textbf{0.073} & 0.11 \\
        D07: verb+able\_reg & \textbf{0.1} & 0.178 \\
        D08: verb+er\_irreg & \textbf{0.061} & 0.114 \\
        D09: verb+tion\_irreg & \textbf{0.061} & 0.082 \\
        D10: verb+ment\_irreg & \textbf{0.039} & 0.065 \\\hline
        \textbf{Lexicography} & $\mathcal{G}$ & $\mathcal{W}$\\\hline
        L01: hypernyms-animals & \textbf{0.159} & 0.212 \\
        L02: hypernyms-misc & \textbf{0.1} & 0.181 \\
        L03: hyponyms-misc & \textbf{0.119} & 0.165 \\
        L04: meronyms-substance & \textbf{0.084} & 0.123 \\
        L05: meronyms-member & \textbf{0.072} & 0.111 \\
        L06: meronyms-part & \textbf{0.123} & 0.166 \\
        L07: synonyms-intensity & \textbf{0.098} & 0.152 \\
        L08: synonyms-exact & \textbf{0.091} & 0.125 \\
        L09: antonyms-gradable & \textbf{0.113} & 0.163 \\
        L10: antonyms-binary & \textbf{0.122} & 0.173 \\\hline
        \textbf{Encyclopedia} & $\mathcal{G}$ & $\mathcal{W}$\\\hline
        E01: country-capital & \textbf{0.051} & 0.067 \\
        E02: country-language & \textbf{0.101} & 0.135 \\
        E03: UK\_city-county & 0.081 & \textbf{0.063} \\
        E04: name-nationality & \textbf{0.06} & 0.062 \\
        E05: name-occupation & \textbf{0.065} & 0.08 \\
        E06: animal-young & \textbf{0.097} & 0.147 \\
        E07: animal-sound & \textbf{0.094} & 0.163 \\
        E08: animal-shelter & \textbf{0.095} & 0.132 \\
        E09: things-color & \textbf{0.073} & 0.11 \\
        E10: male-female & \textbf{0.072} & 0.105 \\
        \hline
    \end{tabular}
    \hfill
    \caption{Comparison of $D$ for $\mathcal{G}$ and $\mathcal{W}$ (lower values mean less dissonance/more bias) over BATS analogies.}
    \label{tab:dissonance}
\end{table}

\newpage

\begin{figure}
  \centering
  \includegraphics[width=0.99\columnwidth]{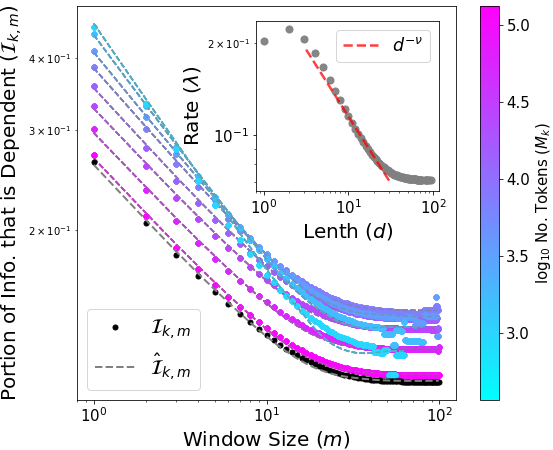}
  \caption{IQR profile for GUM (points), measured on 75 randomizations, scanning \textit{all} values of $m$ and $k$ at powers of two, alongside dependence model fits (dashed lines). Color indicates the $\log$-average number of tokens for each sample size, $k$. Inset shows the whole corpus' (black points/dashed lines) scaling exponent, $\nu$, as a natural fit (red dashed line) for the cumulative-rate of dependencies, $\lambda$, that one can observe against the co-occurrence background (gray points), as the dependence length \textit{and} co-occurrence window size ($d=m$) increase.}
  \label{fig:GUM}
\end{figure}

\section{Quantifying Independence}
\label{sec:independence-appendix}
Here, we first empirically review the IQR's overall shape in the context of the Georgetown University Multilayer (GUM) Corpus~\cite{zeldes2017}, which affords opportunity to model the IQR as a function of the context-window size, $m$, allowing for determination of if and how linguistic dependencies contribute to the statistical dynamics of co-occurrence dependence, as measured by $\mathcal{I}_{k,m}$. 

\subsection{Measuring Dependencies Empirically}
We perform samples amongst GUM's $k_\text{max} = 150$ documents. In Fig.~\ref{fig:GUM}, we observe that even for the highest-IQR ($1$-document) samples, $\mathcal{I}_{1,m}<0.5$, i.e., the IQR's values are less than one half for all window and sample sizes. For now, we'll forego the effects of $k$ and focus on how $\mathcal{I}_{k,m}$ is a decreasing function of $m$. This should be expected, i.e., that lower-$m$ values exhibit less independence, and as we now show, this can be understood according to the dashed-line models that Fig.~\ref{fig:GUM} exhibits. 

\subsection{Modeling the Density of Dependencies}
\label{sec:modeling-dependencies}
In one sense, dependency parsing grammatically determines a rule for `who' each given word's company is. By annotators, each dependency must be determined from the full range of co-occurrences available in the given sentence. As it turns out, dependencies are believed to have their own power-law statistical relationships between dependence length-$d$ (the number of other words up through the dependence) and frequency~\cite{chen2019}. Previously, raw counts of dependencies were observed to form a power-law-like distribution that scaled as $d^{2.5}$. However, we wish to model dependencies against their background of co-occurrences in sentences. This means the sentence-length distribution modulates a critical bound on co-occurrence IQR. Next, we use the nature of how this background of contrastive information in sentences saturates with increasing $m$ to model the lower limit of IQR values observed in Fig.~\ref{fig:GUM}. To produce this model, we first formally state our conjecture, and then derive the model.

\vspace{5pt}\noindent\textbf{Conjecture.} Linguistic dependencies are `the company words keep' from the distributional hypothesis, and underpin the statistical dependencies one can measure against the co-occurrence background via $\mathcal{I}_{k,m}$. We find support for this conjecture by developing a parametric model for $\mathcal{I}_{k,m}$, fitting it over the GUM corpus, and exhibiting how its fit corresponds to the density of linguistic dependencies against their co-occurrence background as the same power-law of $m$.

\subsubsection{Forming Dependence Models}
Define $g_m$ to be the number of dependencies of a given length: $g_m \propto m^{-\nu}$, where $\nu$ is a positive, power-law scaling exponent. Alongside $\nu$, we define a maximum dependence length, $m_\text{max}$, as a model parameter to form the IQR's estimator as a function of the context window, $m$. First, we approximate the $\nu$-power-law's cumulative distribution function over dependencies covered by the window $m$:
$$
G_m = \int_0^{m} g_\ell d\ell \:\:=\:\: \frac{m^{1-\nu}}{m_\text{max}^{1 - \nu}}.
$$
Here, totality requires setting $G_m = 1$ for all $m > m_\text{max}$. This definition for $G_m$ allows us to accurately show how independent statistics saturate co-occurrence models as $m$ becomes large. However, predicting the IQR for empirical co-occurrences depends heavily on the sentence-length distribution, which determines how many co-occurrences exist per each center word. Any sentence of length $L$ will induce $L(L-1)$ co-occurrences if $m \geq L$. If $m < L$, the longer-range co-occurrences are ignored, making the general, total number of co-occurrences per sentence of length $L$ equal to:
$$
T_{L,m} = \left(\min(m, L) - 1\right)\cdot\left(2L - \min(m, L)\right)
$$
Supposing there are $S_L$ sentences for each $L$ and that the longest sentence length is $L_\text{max}$, the total number of co-occurrences in an $m$-radius sliding-window model will be $T_m = \sum_{L = 1}^{m}S_LT_{L,m}$,
which defines the limiting-$m$ co-occurrence model---with the longest range dependencies---by $T_{L_\text{max}}$. Setting $T_0 = 0$ and denoting the total unigram frequency in a sample of size $k$ as $M_k$ allows us to define the co-occurrence sampling rate for all $m = 1, \cdots, L_\text{max}$ as:
$$
q_m = \frac{T_m - T_{m-1}}{2M_k}
$$
Given any $m$, $G_m$, represents the cumulative portion of all dependences sampled from a $2m$-window, according to our base model. This base model is then modulated by the sampling rate, $q_m$, and scaled by a constant $\rho$:
$$
\hat{\mathcal{I}}_{k,m} = \rho\cdot\left[q_m\frac{G_m}{2m} + (1 - q_m)\frac{M_k}{T_m}\right]
$$
which parametrically defines the average number of dependencies word. Overall, this formulation can intuitively by understood to transfer---via the sampling rate---power-law varying dependence density smoothly, into a limiting $\rho$ number of dependences per word in the sample: $\frac{M_k}{T_{L_m}}$. Via $\rho$, the model assumes on average that each word depends on $\rho$ of the other words in the same sentence (co-occurrences), meaning $\rho\in[0,\frac{T_{L_\text{max}}}{M}]$.

\begin{figure*}
  \centering
  \includegraphics[width=0.99\columnwidth]{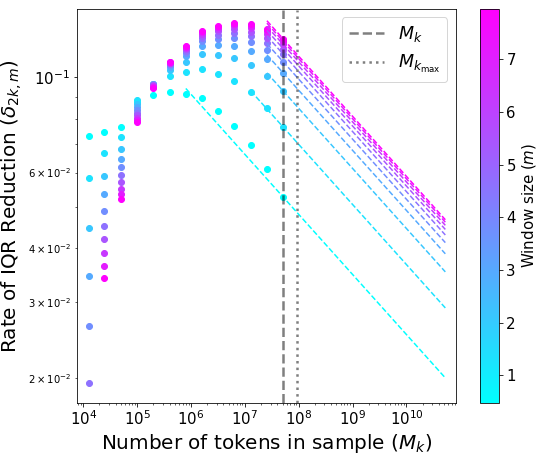}
  \includegraphics[width=0.99\columnwidth]{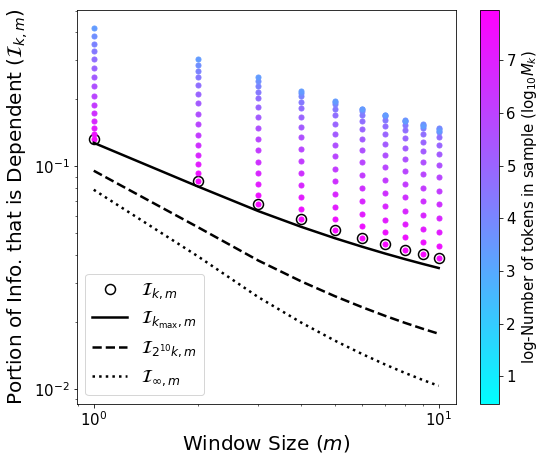}
  \caption{(Left) Rate of IQR reduction presented against the number of tokens in 75 samples of of size $k$ taken at powers of two from the training articles of the WikiText-103 corpus for context-window sizes $m$ up to $10$ (indicated by color). Past the size of the whole corpus (black dotted line), the reduction rates are extrapolated with a power-law to model limiting behavior (spectrum of dashed lines). (Right) IQR profile for the WikiText-103 training corpus up to the largest two-power sample size (most pink) for context-window sizes up to $m=10$. Past the largest-sampled size (circled points), the rate-reduction power law is used to extrapolate (e.g., the black dashed line) beyond the corpus (black dashed line) to compute non-zero limiting IQR values across $m$-window sizes.}
  \label{fig:WikiText}
\end{figure*}

\subsection{Fitting Dependence Models}
As can be seen in Fig.~\ref{fig:GUM}, $\hat{\mathcal{I}}_{k,m}$ can be parameterized to fit the IQR quite strongly. Compared to the scale of co-occurrences absorbed by large representation models, the GUM data set is quite small. However, it affords a critical opportunity to observe how the power-law exponent, $\nu$, corresponds to the density of \textit{linguistic} dependencies against the co-occurrence background. Critically, we find in Fig.~\ref{fig:GUM}'s inset a correspondence between $\nu$ and the rate of decay of linguistic dependencies against the co-occurrence background for full order of magnitude in the dependence-length distribution. While the scale of these results are small inside of GUM, they strongly support our conjecture. However, a number of questions and challenges emerge from these experiments. Practically, sampling from GUM has exhibited how the longest-range dependencies are simply not available to model in shorter sentences---the effects of small samples and varying sentence lengths can be seen in the empirical roughness for larger-$m$ windows in smaller-$k$ samples. Furthermore, larger samples pose combinatoral scaling for larger-$m$ co-occurrences that makes direct measurement of the IQR prohibitive. To compound these issues, we seek to know if any limiting bounds on the IQR exist, i.e., we ask: ``would the IQR from an infinite number of documents be zero?''

\subsection{Bounding Dependence from Below}
\label{sec:IQR-bound}
To see the effects of scale on the IQR, more data are required for experimentation than are available within the GUM corpus. Even if we can't expect linguistic dependencies to be annotated, we still wish to control for tokenization. Hence, to bound the IQR we work next with the well-known WikiText language modeling benchmark~\cite{merity2016}, which was expanded in v103 to over $30,000$ documents (roughly $200$-times the size of GUM). Our objective in this section is to bound the IQR, and large-$m$ measurement becomes intractable for large $k$-document samples. Hence, we will restrict $m\leq 10$ for all bounding experiments. 

Suppose we fix any window size, $m$, let $k$ be the number of documents in a sample, and $\mathcal{I}_{k,m}$ denote the IQR's average value. For any sample size, $k$, then let $M_k$ denote the expected number of tokens in the sample, and observe that doubling $k$ will double the expected number of tokens: $M_{2k} = 2M_k$. We're most interested in \textit{the rate of IQR reduction incurred from doubling the number of sampled documents}, which we denote by $\delta_{2k, m}$. Averaging across samples, we have generally observed the IQR to fall into a well-ordered---perhaps power-law---pattern of decay (Fig.~\ref{fig:WikiText}, left). This means that for large values, $k_1 \leq k_2$, one can expect: $\delta_{2k_2, m} \leq \delta_{2k_1, m}$. Next, we prove that these observed conditions result in the existence of a positive lower bound for the IQR, which exists below the IQR one could measure for \textit{any} document-sample size, $k$.

\vspace{5pt}\noindent\textbf{Theorem.} If the $k$-limiting behavior of the IQR-reduction rate is power-law decay: $\delta_{2k, m}\propto M_k^\gamma$, the IQR's limiting, $\mathcal{I}_{\infty,m}$, values are positive.

\vspace{5pt}\noindent\textbf{Proof.} Supposing we fix any window size, $m$, let $k$ be the number of documents in a sample, and $\mathcal{I}_{k,m}$ denote the IQR's average value. For any sample size, $k$, let $M_k$ denote the expected number of tokens in the sample, and observe that doubling $k$ will double the expected number of tokens: $M_{2k} = 2M_k$. We next express the rate of IQR reduction incurred from doubling the number of sampled documents by $\delta_{2k, m}$. Averaging across samples, we have generally observed the IQR to fall into a well-ordered---perhaps power-law---pattern of decay. This means that for large values, $k_1 \leq k_2$, one can expect: $\delta_{2k_2, m} \leq \delta_{2k_1, m}$. 

We wish to know about any limiting dynamics of $\mathcal{I}_{k,m}$ for large $k$, which under the observed pattern of decay ammounts to asking if the IQR converges to zero or a positive limit. In either case, we'll refer to any limiting quantity as ${I}_{\infty, m}$, which describes the portion of information that is dependent in a co-occurrence model of context for a \textit{population} of data, i.e., an arbitrarily-large sample. Critically, this limit expresses the dependence in \textit{how} a population of language was used, separately from the dependence on \textit{what} its samples convey.

Assuming $k$ is large enough to well order the reduction rate at a window size of $m$, we use $\delta\mathcal{I}_{2k, m}$ to write an IQR-update rule for doubled samples:
$$
\mathcal{I}_{2k, m} = \mathcal{I}_{k, m}(1 - \delta_{2k, m}),
$$
Applying recursion over this equation allows us to express IQR values for arbitrarily large samples:
$$
\mathcal{I}_{2^nk,m} =
\mathcal{I}_{k,m}
\prod_{l=1}^n(1 - \delta_{2^lk, m})
$$
However, to study a limiting value for the IQR we apply the reduction rate in series: $\mathcal{I}_{\infty, m} = \mathcal{I}_{k, m} - \sum_{n=0}^\infty
\mathcal{I}_{2^{n}k,m}\delta_{2^{n+1}k,m}$.
With this, we can substitute the product form for $\mathcal{I}_{2^nk,m}$ into our expression for $\mathcal{I}_{\infty, m}$ to produce:
$$
\mathcal{I}_{\infty, m} = \mathcal{I}_{k, m}\left[1 - \sum_{n=0}^\infty
\delta_{2^{n+1}k,m}\prod_{l=1}^n(1 - \delta_{2^lk, m})\right]
$$
Decreasing monotonicity in the reduction rate implies that the fastest-decaying extreme occurs when the reduction rate is a constant. Supposing this to be the case, we assume a critical sample size, $k_m$, past which a constant, $\delta_m$, describes the reduction rate. When one substitutes this into our expression for the IQR's limit, a geomtric series emerges which unsurprisingly brings the IQR's limit to its low (0-valued) extreme:
$$
\mathcal{I}_{\infty, m} = \mathcal{I}_{k_m, m}\left[1 - \delta_m\sum_{n=0}^\infty
(1 - \delta_m)^n\right] = 0
$$
One can in fact approximate the reduction rate empirically by computing a quotient of expected IQR values from samples of documents:
$$
\delta_{2k, m}\approx\frac{E\left[\mathcal{I}_{k,m} - \mathcal{I}_{2k,m}\right]}{E\left[\mathcal{I}_{k,m}\right]}
$$
When measured, we find that for $k$-samples larger than some critical sample size, $k_m$ (dependent on $m$), the reduction rate appears to scale like a power law in the number of tokens sampled:
$$
\delta_{2k, m} \approx \frac{M_{k}^{-\gamma}}{10^{b_m}}
$$
This models $\delta_{2k, m}$ with a scaling exponent, $\gamma$, and constant of proportionality, $10^{b_m}$, the latter of which is dependent upon the window size, $m$. Utilizing this empirically-motivated power-law, we obtain a different form for the limiting IQR: 
$$
\mathcal{I}_{\infty, m} = \mathcal{I}_{k, m}\left[1 - \sum_{n=0}^\infty
\frac{M_{2^{n+1}k}^{-\gamma}}{10^{b_m}}\prod_{l=1}^n\left(1 - \frac{M_{2^lk}^{-\gamma}}{10^{b_m}}\right)\right] 
$$


To bound $\mathcal{I}_{\infty, m}$ from below one can replace each of the products of $n$ with unity. The identity: $M_{2k} = 2M_{k}$ allows further generalization of the doubling numbers as: $M_{2^{n+1}k} = 2M_{2^nk}$, whose substitution into the bound produces a convenient form and infinite geometric series of ratio $2^{-\gamma}$:
\begin{equation}
\mathcal{I}_{k, m}\left[1 - 
\frac{M_{2k}^{-\gamma}}{10^{b_m}(1 - 2^{-\gamma})}
\right] < 
\mathcal{I}_{\infty, m}
\label{eq:IQR-bound}
\end{equation}
To assure this lower bound is positive we now need only require:
$$
\frac{M_{2k}^{-\gamma}}{10^{b_m}(1 - 2^{-\gamma})} < 1
$$
which amounts to asserting that the average number of tokens, $M_{k}$, from the original sample size of $k$ is sufficiently large to bound the the a positive-valued function parameterized by $m$ via $\gamma$ and $b_m$ that notably has no dependence on $k$:
$$
10^{-b_m/\gamma}(2^{\gamma} - 1)^{-1/\gamma} < M_{k}
$$
Hence, positive lower bounds on the IQR could be confirmed experimentally for any given $m$ by increasing the initial sample size of an the analysis. This likewise provides a means for the observing the elimination of transient independence in a co-occurrence model, where specifically, as $k$ is increased, the positive, lower bound tightens to the limiting IQR value from below. $\blacksquare$

\subsubsection{Computing a Bound for Dependence}
Returning to the WikiText Corpus, we repeatedly sample the available powers of $k=2^n$, which for WikiText-103 allows $n\leq14$, since the total number of documents in the collection is roughly: $k_{\text{max}} = 2^{15}$. These doubling samples are used to empirically compute our approximations of the reduction rate's average behavior in Fig.~\ref{fig:WikiText} (left), and the value of $\gamma$ is optimized over only those values for this the power-law decay is apparent. Moreover, we constrain the values of $b$ to satisfy a continuous model projecting from the data, i.e., optimization is only performed over $\gamma$. Once $\gamma$ and the different values of $b_m$ are established, $2^{10}$ iterative updates to $\mathcal{I}_{k, m}$ for $2^{15}$ to produce the large-$k$ IQR values needed for stable computation the IQR's limit, $\mathcal{I}_{\infty,m}$. Each of these modeling components used in computing the IQR's bound is exhibited in Fig.~\ref{fig:WikiText} (right), and confirm the nature of our bounding result.

\end{document}